\documentclass{fairmeta}
\usepackage{mathptmx}% Times-like font
\usepackage{helvet}     % Helvetica
\usepackage{courier} 
\usepackage{xcolor}
\usepackage{enumitem}
\usepackage{float}
\usepackage{xltabular}
\usepackage{pgfplots}
\pgfplotsset{compat=1.18}
\usepackage{wrapfig}
\usepackage{tcolorbox}
\usepackage{listings} % Added for lstdefinestyle

\usepackage{microtype}  % Better typography and spacing
\usepackage{url}        % Better URL handling
\usepackage[raggedrightboxes]{ragged2e}
\usepackage{caption}

\newcommand{\agriname}{Krishi Sathi} 

\definecolor{lightgray}{gray}{0.95}
\lstdefinestyle{chatml}{
    backgroundcolor=\color{lightgray},
    basicstyle=\ttfamily\small,  % Made font smaller
    frame=none,
    columns=fullflexible,
    breaklines=true,
    keepspaces=true,
    showstringspaces=false,
    breakatwhitespace=true,      % Break at whitespace
    postbreak=\mbox{\textcolor{red}{$\hookrightarrow$}\space}, % Show line breaks
}

\title{Intent Aware Context Retrieval for Multi-Turn Agricultural Question Answering}

\author{Abhay Vijayvargia}
\author{Ajay Nagpal}
\author{Kundeshwar Pundalik}
\author{Smita Gautam}
\author{Atharva Savarkar}
\author{Viraj Thakur}
\author{Pankaj Singh}
\author{Rohit Saluja}
\author{Ganesh Ramakrishnan}

\affiliation{BharatGen Team}

\abstract{\justifying Indian farmers often lack timely, accessible, and language friendly agricultural advice, especially in rural areas with low literacy. To address this gap in accessibility, this paper presents a novel AI-powered agricultural chatbot, \agriname\footnote{Krishi Sathi can be accessed at: \href{https://bharatgen.com/products/krishi-saathi}{Krishi Sathi}}, designed to support Indian farmers by providing personalized, easy-to-understand answers to their queries through both text and speech. The system’s intelligence stems from an IFT model \cite{wei2021finetuned}, subsequently refined through fine tuning on Indian agricultural knowledge across 3 curated datasets. Unlike traditional chatbots that respond to one-off questions, Krishi Sathi follows a structured, \emph{multi-turn conversation flow} to gradually collect the necessary details from the farmer, ensuring the query is fully understood before generating a response. Once the intent and context are extracted, the system performs Retrieval-Augmented Generation (RAG) by first fetching information from a curated agricultural database and then generating a tailored response using the IFT model. The chatbot supports both English and Hindi languages, with speech input and output features (via ASR and TTS) to make it accessible for users with low literacy or limited digital skills. This work demonstrates how combining intent-driven dialogue flows, instruction-tuned models, and retrieval-based generation can improve the quality and accessibility of digital agricultural support in India.

This approach yielded strong results, with the system achieving a Query Response Accuracy of 97.53\%, 91.35\% contextual relevance and personalization, and a query completion rate of 97.53\%. The average response time remained under 6 seconds, ensuring timely support for users across both English and Hindi interactions. \footnote{Base model can be accessed at: \href{https://aikosh.indiaai.gov.in/home/models/details/bharatgen_param_1_indic_scale_bilingual_foundation_model.html}{BharatGen Param-1 (Base Model)}}
.}

\date{\today}
\correspondence{kundeshwar.pundalik@tihiitb.org}

\begin{document}

\twocolumn[
\maketitle
\vspace{0.5cm}
]
\section{Introduction}
India's agricultural sector involves millions of farmers who often need timely advice on crops, weather, pests, pricing, and government schemes. However, access to reliable information is still a challenge, especially in rural areas where literacy may be limited. With the growth of mobile connectivity and voice-based interfaces even in rural areas, conversational AI offers a new way to bring relevant, real-time support to farmers.

This paper introduces an intelligent agricultural chatbot, ‘Krishi Sathi’, which helps bridge this gap using recent advances in language modeling and retrieval. At the heart of the system is an Instruction-Fine-Tuned \cite{wang2023selfinstructaligninglanguagemodels} language model, trained on agricultural datasets specific to the Indian context. These include knowledge from ICAR\footnote{\url{https://icar.org.in}} (Indian Council of Agricultural Research), Vikaspedia\footnote{\url{https://www.vikaspedia.in}}, and other government advisories. Krishi Sathi uses a \emph{guided conversation flow} to extract user intent and context in a step-by-step manner. This ensures even vague or incomplete initial queries are clarified before a response is generated.

Once enough information is gathered, the system employs a Retrieval-Augmented Generation (RAG) proposed by \cite{lewis2021retrievalaugmentedgenerationknowledgeintensivenlp} using Dense Retrieval \cite{DBLP:journals/corr/abs-2004-04906} techniques to search a backend agricultural database for relevant content. The IFT model then uses this retrieved content to craft a response that is accurate, relevant, and, most importantly, reliable.

The chatbot currently supports English and Hindi, and the architecture allows for easy extension to other Indian languages in the future. To improve accessibility, especially for farmers who prefer speaking over typing and reading, the system also includes speech-to-text (ASR) for input and text-to-speech (TTS) for output.

The key contributions of this work are twofold: (1) a multilingual, speech-enabled, intent-aware chatbot for low-literacy users in agriculture, and (2) a retrieval-augmented pipeline fine-tuned on curated Indian agricultural datasets for accurate and contextual response generation.

The paper is structured as follows. Section 2 covers existing work in agricultural chatbots and the concepts explored and utilized while implementing Krishi Sathi. Section 3 provides details of the data curation process as well as the key components of our system - RAG module, LLM inference, and the IFT model. Section 4 gives a detailed description of the workflow when a query is being processed. The subsequent sections cover discussion, results, and plans for future work.

\section{Related Work}
The past decade has witnessed a rapid evolution in agricultural decision support systems, moving from early rule based advisories and static mobile platforms to data driven machine learning solutions capable of delivering personalized recommendations. Comprehensive surveys, such as  \cite{liakos2018}, underscore how big data techniques, from sensor networks to satellite imagery, can optimize tasks like yield prediction, pest detection, and resource management. Likewise,  literature survey by \cite{das2018} categorizes key AI technologies, including expert systems, neural networks, and computer vision, and highlights their strengths and limitations in areas such as soil management and disease diagnostics. More recent works, such as \cite{javaid2023}, identify the transformative potential of AI to drive decision support to deliver real time, location specific advisories, addressing the lack of adaptability and personalization in the early rule based approaches.

In the Indian context, government initiatives like Kisan Call Centers (KCC) \footnote{\url{https://www.manage.gov.in/kcc/kcc.asp9}}, launched in 2004 as a toll free helpline for farmers in 22 local languages, sought to bridge information gaps by connecting farmers directly to agricultural experts. Despite its broad approach, the system faced several limitations that affected its overall ability to help farmers. As detailed in the study by \cite{grover2017}, the effectiveness of KCC was constrained by challenges such as inconsistent response quality, lack of localized experts, and difficulties in handling the volume of queries. The study further details that in states like Gujarat, the percentage of calls answered effectively was only 45.7\%, as compared to 93.9\% in Punjab, which may be the result of uneven distribution of agricultural experts who speak local languages. Furthermore, according to the survey findings in the report, only 57\% of farmers reported that the advice provided by KCC actually solved their problem, meaning 43\% of the farmers still faced problems even after the consultation by KCC experts. The satisfaction number dips even further for queries related to market and government schemes. Even though existing systems such as KCC and the mKisan \footnote{\url{https://mkisan.gov.in}} portal are working towards helping farmers, there is still a need for a sophisticated system that can cater to a larger population of farmers with a higher rate of satisfaction. 

Such sophisticated systems have already been introduced in small scale experiments, and have shown positive results. As stated in \cite{agripilot2025}, in Baramati, Maharashtra, around 1,000 farmers participating in a pilot initiative led by the Agriculture Development Trust (ADT) \footnote{\url{https://www.kvkbaramati.com/agridevelop.aspx}}, in collaboration with Microsoft and Agripilot.ai \footnote{\url{https://agripilot.ai}}, reported enhanced sugarcane growth with stalks and more sucrose yield, alongside reduced input usage. One of the biggest factors in this achievement is making available the Agripilot.ai app available in Marathi, which enabled even local farmers to use the agent. Similarly, according to \cite{saagu2024}, Telangana’s Saagu Baagu initiative helped 7000 chili farmers get access to AI based advisories. This led to a notable improvement in yield per acre and an increase in unit prices, along with a reduction in pesticide use.

Recent advancements in AI driven agricultural chatbots, similar to the successful small scale experiments, have significantly enhanced information accessibility among Indian farmers. Farmchat \cite{jain2018farmchat} introduced a conversational agent tailored to address the information needs of farmers in rural India, emphasizing localized and personalized responses. Krushi - the Farmer Chatbot \cite{momaya2021} aimed to assist farmers by leveraging predefined responses to answer queries related to crops, fertilizers, and weather. AgriLLM \cite{didwania2024} leveraged a transformer based language model trained on approximately 4 million real world farmer queries from Tamil Nadu, showcasing the potential of LLMs in automating query resolution for farmers.

Despite their advancements, these systems exhibit several limitations that can be addressed through the application of more sophisticated technologies. FarmChat’s pipeline is tightly coupled to a single crop—potato farming—relying on a manually curated knowledge base and rule‑based dialogue tailored specifically to that crop. As a result, adding support for other crops requires significant re-engineering of the system. To address the challenge of retaining specialized information across multiple domains, the proposed system incorporates a Retrieval Augmented Generation (RAG) framework. This enables dynamic access to a structured, domain-specific knowledge base containing region and crop specific data. To enhance retrieval within RAG, the system draws upon Dense Passage Retrieval techniques proposed by \cite{sachan2023improving}, which significantly improve retrieval performance in open domain question answering by encoding both query and passage into dense representations.

Beyond retrieval based limitations, systems like Krushi rely heavily on static responses and lack the flexibility to handle nuanced and evolving queries. Similarly, AgriLLM, though trained on millions of real world farmer queries, processes all queries uniformly without understanding specific user goals, which might result in general or misaligned responses for specialized needs. To overcome these shortcomings, the proposed solution adopts Task Oriented Dialogue with In Context Learning presented by \cite{bocklisch2024task}, allowing multi turn interactions centered around user defined goals such as pest identification, fertilizer suggestions, or weather guidance. This is paired with intent detection and slot filling models as discussed in \cite{weld2021surveyjointintentdetection}, ensuring precise interpretation of user queries. To further enhance adaptability in low resource scenarios, the system integrates In Field Tuning \cite{han2024parameter} and In Context Learning \cite{Jeong_2024}, which aligns the model with agricultural task specific prompting while giving few shot examples for enhanced responses. Together, these techniques create a context aware assistant capable of addressing complex and diverse queries more effectively than prior proposed systems.

\section{Methodology}
This section outlines the methodology adopted to develop a domain specific instruction dataset and inference pipeline tailored for the agriculture sector, specifically focused on grapes and onions. The process is divided into several key stages as shown in Figure \ref{fig:pipeline}: data collection, curation, and extraction from diverse agricultural sources; classification of collected data into distinct categories; and the generation of instruction-based datasets aligned with those categories. Subsequently, general-purpose large language models are leveraged and evaluated for their performance, followed by the design and integration of a retrieval module to enhance context-aware responses. Finally, the model undergoes in field tuning using real world queries to ensure practical applicability and domain relevance. The following subsections describe these stages in detail.

\subsection{Data Collection}
The data collection process was initiated with the identification of credible and domain-specific data sources. The finalization of sources was conducted in consultation with 3 agricultural experts, who assisted in assessing domain relevance, filtering out low-quality or irrelevant content, and validating the authenticity of the sources. The expert-guided approach ensured that the dataset was aligned with the specific needs and complexities of the domain. Key data sources included:

\begin{figure}[h]
    \centering
    \includegraphics[width=0.9\linewidth, keepaspectratio]{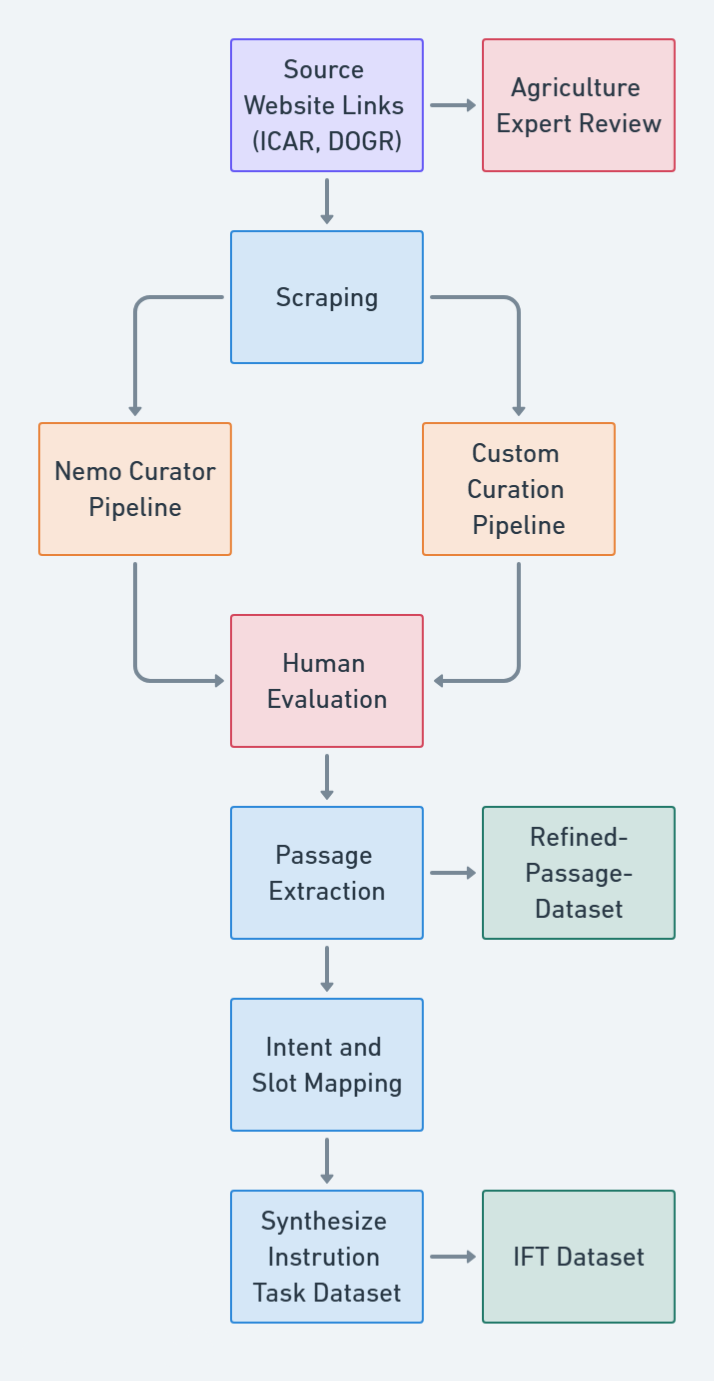}
    \caption{Data Curation Pipeline}
    \label{fig:pipeline}
\end{figure}

\begin{enumerate}
    \item The Indian Council of Agricultural Research (ICAR), the apex body responsible for coordinating agricultural research and education in India. \cite{icar2024}
    \item Vikaspedia, a government-managed multilingual knowledge portal developed under the National e-Governance Plan. \cite{vikaspedia2024}
    \item ICAR-affiliated institutes such as the National Research Centre for Grapes (NRCG) and the Directorate of Onion and Garlic Research (DOGR), both of which conduct strategic and applied research in horticultural science.\cite{nrcg2024,dogr2024}
\end{enumerate}

Once the sources were finalized, a systematic web scraping process was undertaken. Publicly accessible content was scraped using tools such as BeautifulSoup while adhering to each website’s terms of use to ensure ethical data collection. The scraping resulted in a raw textual dataset comprising approximately 20.4 million tokens.

The scraped data primarily consisted of unstructured textual information, including articles, technical guidelines, tables, and research summaries, which were subsequently passed to the preprocessing pipeline described in the next section.

\subsection{Data Curation and Extraction}

\begin{tcolorbox}[colback=gray!10, colframe=black!50, boxrule=0.5pt, arc=4pt, left=6pt, right=6pt, top=6pt, bottom=6pt]
\textbf{Curated Passage Example:}

"In order to address the issues in crop growth and getting high yields, several actions should be taken. Firstly, it is important to promote crop and cropping systems that are suitable for the specific agro-climatic conditions. This can help to ensure that crops are able to thrive and produce high yields. Secondly, it is recommended to diversify crops and cropping systems by incorporating livestock, fisheries, horticulture, dairy, and agroforestry. This can help to improve the overall sustainability and resilience of the agricultural system, as well as provide a more diverse range of products and income sources.Thirdly, steps should be taken to create sources of protective irrigation through the construction of check dams, tanks, farm ponds, shallow or medium tube wells, and dug wells. This can help to ensure that crops have access to the water they need, even during periods of drought or low rainfall.Finally, it is important to adopt technologies that can improve water use and moisture conservation. This can include the use of efficient water application systems, land leveling, field bunding, contour bunding, trenches, and mulching. The ridge and furrow method can also be effective for conserving moisture and improving water use efficiency. By implementing these measures, it is possible to significantly improve the productivity and sustainability of agricultural systems."
\end{tcolorbox}
From the initial raw corpus of 20.4 million tokens, a two-week curation process yielded a refined dataset of approximately 12 million tokens, retaining nearly 59\% of the original data. A hybrid approach combining automated preprocessing and human evaluation was employed to ensure both syntactic quality and domain relevance. A team of 3 domain experts periodically reviewed samples from different pipeline stages to validate quality and relevance.

Tools, including Nemo Curator \footnote{\url{https://developer.nvidia.com/nemo-curator}} and custom-made curation pipelines, were implemented to perform two major tasks for preprocessing: modification and filtering. The modification stage included boilerplate removal, HTML tag stripping, Unicode reformatting, and white space removal. Following modification, the data was passed through a series of filters, including a word count filter, a whitespace filter, a repeating n-gram filter, and a numerical dominance filter to eliminate noisy or low-quality text retained even after the modification pipeline was implemented.

Finally, high-quality segments from the curated data were extracted as coherent and context-rich paragraphs using the Mistral-Nemo-Instruct-2407 \footnote{\url{https://huggingface.co/mistralai/Mistral-Nemo-Instruct-2407}} model. This dataset, referred to as the Refined-Passage-Dataset, comprises 150,000 domain refined and expert reviewed passages.

\begin{table}[ht]
  \centering
  \resizebox{\columnwidth}{!}{
    \begin{tabular}{|p{4cm}|p{4cm}|}
      \hline
      \textbf{Intent} & \textbf{Slots} \\
      \hline
      \multirow{4}{*}{Vineyard Variety Selection} 
        & Grape Variety \\
        & Climate \\
        & Expected Yield Potential \\
        & Soil Type \\
      \hline
      \multirow{4}{*}{Irrigation Management} 
        & State \\
        & Season \\
        & Seed Variety \\
        & Soil Testing \\
      \hline
      \multirow{3}{*}& Fertilizer Type \\
        {Fertilization and Nutrient }& Fertilization Schedule \\
        {Management}& Micronutrient Deficiency in Soil \\
      \hline
    \end{tabular}
  }
  \caption{\label{intent-slot-grapes}
    Intent Slot Mapping for Grapes
  }
\end{table}

\begin{table}[ht]
  \centering
  \resizebox{\columnwidth}{!}{
    \begin{tabular}{|p{4cm}|p{4cm}|}
      \hline
      \textbf{Intent} & \textbf{Slots} \\
      \hline
      \multirow{4}{*}{Time of Transplanting} 
        & State \\
        & Season \\
        & Seed Variety \\
        & Time of Sowing \\
      \hline
      \multirow{5}{*}& State \\
        {Integrated Pest Management }& Season \\
        {Protocols}& Seed Variety \\
        & Time of Sowing \\
        & Time of Transplanting \\
      \hline
      \multirow{2}{*}{Land Clearing \& Tilling} 
        & Fertilizer Type \\
        & Fertilization Schedule \\
      \hline
    \end{tabular}
  }
  \caption{\label{intent-slot-onions}
    Intent Slot Mapping for Onions
  }
\end{table}

\subsection{Data Classification and Instruction Dataset Generation}
After the curation process, the next phase focused on classifying the obtained 12 million tokens into intents and slots, and generating an instruction based dataset for fine-tuning. The first step involved defining and annotating intents for 2 agricultural crops: grapes and onions. Specifically, 25 distinct intents were defined for grapes and 22 for onions, each intent comprising 2 to 5 slots to capture domain specific information. In this context, an intent is referred to as a high level goal, as seen in Table 1. A total of approximately 18000 annotated examples were collected across both domains. The annotation process was conducted by a dedicated team of 3 trained annotators, who manually reviewed and labeled the data, ensuring high quality semantic alignment with domain-specific tasks. 

For constructing the instruction based dataset, the Refined-Passage-Dataset was utilized. The passages were used to create 15 conversation based instruction tasks that simulate realistic agricultural scenarios, facilitating the fine tuning of the LLM used for generating the final output. These tasks enable the model to learn and respond effectively to domain specific instructions. The tasks generated using the passages were then reviewed by the 3 domain experts to ensure consistency with the intended classification and broader domain context. The final output was a structured instruction based dataset formatted as shown in Example 2.

\subsection{General LLM Inference Functions}
Considering the extensive decision making required throughout the pipeline, the Param-1-2.9B-instruct is employed for all general purpose functions. It supports key system components, including (1) the router (routing queries between general purpose and domain specific models), (2) the crop classifier, (3) the intent classifier, and (4) the question generator for refining intent slot mappings. Additionally, it handles general purpose queries and casual user interactions, preserving the fine tuned domain specific model for agricultural queries. This architectural separation optimizes latency and resource utilization.

According to Bharatgen, Param-1-2.9B outperforms comparable models such as Sarvam-1 (2B) across several benchmarks. It achieved 73.4\% on HellaSwag \cite{hellaswag}, 61.6\% on Winogrande \cite{winogrande}, 79.3\% on PIQA \cite{piqa},  46.0\% on MMLU \cite{mmlu}, 36.1\% on MMLU Hindi, 48.3\% on MILU Hindi, and 49.7\% on MILU English \cite{milu}. These performance and architectural characteristics, including improved instruction following and few-shot reasoning, while still functioning on only 2.6 billion parameters, justify its role as the backbone for non specialized tasks within the proposed pipeline.
\subsection{Retrieval Module}
To supplement the language generation with grounded knowledge, a dense vector retrieval module was integrated into the pipeline. The primary motivation was to ensure that the model’s responses were factually aligned with domain specific information, especially when answering agricultural queries that require high precision. We formulate this as a nearest neighbor problem in a high dimensional embedding space, where the goal is to retrieve the most relevant document corresponding to a query vector. The retriever module was built using the all-mpnet-base-v2 \cite{mpnet} model from the Sentence Transformer library \footnote{\url{https://huggingface.co/sentence-transformers/all-mpnet-base-v2}} and deployed on a CUDA-enabled H100 GPU for efficient embedding throughput. Documents were preprocessed by removing metadata and encoded into 768-dimensional vectors. Approximately 150,000 embeddings were generated and indexed using the Qdrant \footnote{\url{https://qdrant.tech}} vector database, which serves as the retrieval backend. All embeddings were computed offline to minimize inference-time latency. During inference, enriched user queries are embedded using the same model and then matched using top-1 nearest neighbour search (k=1).

\subsection{In Field Tuning for Domain Expertise}
The Param-1-2.9B model was fine-tuned (supervised) on the curated agricultural corpus described in the previous subsections using NVIDIA NeMo's Megatron GPT SFT~\footnote{\url{https://docs.nvidia.com/nemo-framework/}} framework on a single-node GPU cluster (8~NVIDIA H200 GPUs) in \texttt{bf16} precision. Training was conducted for 3 epochs with a global batch size of 1024, achieved via a micro-batch size of 4 per GPU and gradient accumulation of 32, and a maximum sequence length of 2048 tokens. The Distributed Fused Adam optimizer (\(\beta_{1} = 0.9\), \(\beta_{2} = 0.98\), weight decay = 0.01) was used in conjunction with a Linear learning rate scheduler configured with 1200 warm up steps and 12000 total steps. The learning rate linearly increased from \texttt{0} to a peak of \(\texttt{5} \times 10^{-6}\), then decayed back to \texttt{0}. No gradient clipping was applied during training. Checkpoints were saved every 1000 steps, retaining the top 20 models based on validation loss, with each generating an inference-ready \texttt{.nemo} file. In production, domain-specific agricultural queries are handled by this fine-tuned model, whereas low-confidence or open-domain queries are routed to the base Param-1-2.9B model.

\begin{tcolorbox}[colback=gray!10, colframe=black!50, boxrule=0.5pt, arc=4pt, left=6pt, right=6pt, top=6pt, bottom=6pt]
\textbf{Instruction-Answer Pair Example:}

\{"instruction": "What measures can be taken to promote crop growth and high yields based on specific agro-climatic conditions?",
"answer": "Based on the passage, one measure that can be taken to promote crop growth and high yields based on specific agro-climatic conditions is to promote crop and cropping systems that are suitable for those conditions. This can help ensure that crops are able to thrive and produce high yields, as they will be planted in conditions that are ideal for their growth. By carefully selecting crops and cropping systems based on the specific agro-climatic conditions, it is possible to optimize crop growth and yield, and make the most of the available land and resources."\}
\end{tcolorbox}

\section{Workflow}
\begin{figure*}[t]
    \centering
    \includegraphics[width=\textwidth, keepaspectratio]{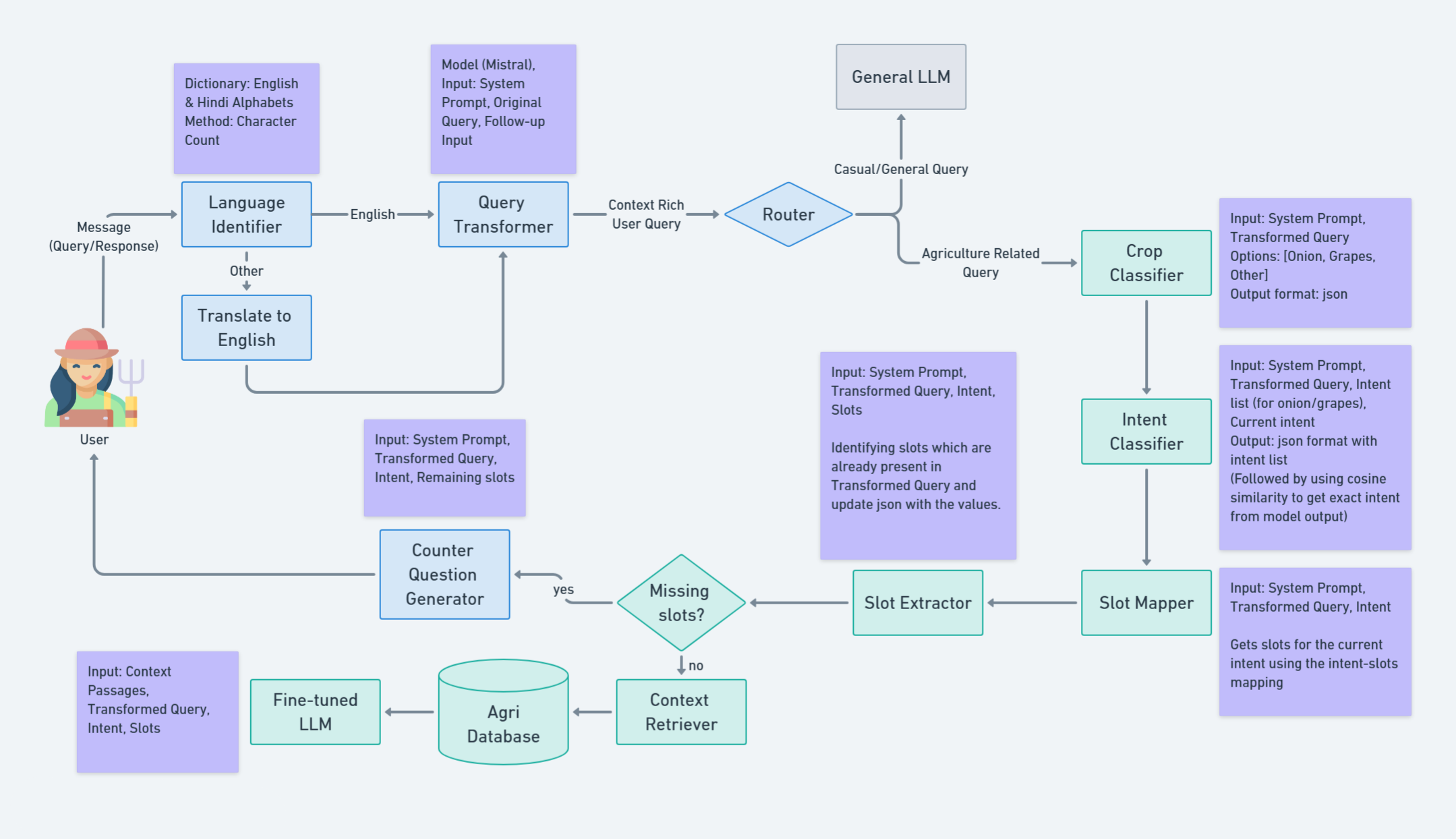}
    \caption{Workflow of Krishi Sathi}
    \label{fig:example_image}
\end{figure*}

This section outlines the detailed workflow of the proposed advisory system. It describes the steps followed by the pipeline once a query is given input by the user, including how the query is processed, contextualized and passed through various modules too generate a tailored, context aware response.

\subsection{Input Transformation, Language Detection, and Translation}
The pipeline is initiated by a user query, which may be provided in text or speech format, as specified by the user. All speech inputs are provided as 16kHz .wav files and are first converted into text using an in house speech-to-text model with 120M parameters deployed on 1 H100 GPU. Once the input is in text format, language identification is performed using a rule based method that compares character frequencies against two custom built dictionaries: one for English and Hindi, and evaluates the query language. For Hindi queries, the Google Cloud Translate \footnote{\url{https://cloud.google.com/translate}} is invoked to convert the input into English. The translation model was chosen due to its robust performance across indic languages and its seamless API integration capabilities.

To ensure consistency across languages, all user queries are mapped to their English equivalents. The pipeline executes exclusively on English text, enabling a unified and optimized processing pipeline. Multilingual support is achieved through the translation model mentioned above, along with an in house TTS model for Hindi speech synthesis. The TTS component is based on F5TTS-small, a 150M parameter model trained on Hindi data and deployed on a single H100 GPU to produce natural and fluent Hindi speech output. 

\subsection{Query Classification and Domain Routing}
Following the translation, the standardized query is routed to the Classification Module, which determines the category of the query among 3 predefined types: Domain Specific, General Knowledge, and Casual Queries. As mentioned in Section 3.4, the classification is performed using the Mistral-Nemo-Instruct-2407 model, a 12B parameter instruction-tuned LLM, selected for its strong few shot generalization and multilingual reasoning capabilities. This model acts as the router and classifier for multiple downstream tasks in the pipeline. 

If the query is classified as Domain Specific, it undergoes crop classification and Intent classification, both implemented using the same Mistral model. These components extract fine grained contextual information such as crop type, region, and season to enable highly targeted information retrieval. The crop specialization is currently limited to two crops: onions and grapes, with 22 and 25 domain specific intent classes each, respectively. Detected slot values are used to augment the query for better grounding in the subsequent retrieval stage.

For General Knowledge and Casual Queries, the system bypasses domain specific classification entirely and forwards the query directly to the Mistral Model for inference. 

\begin{tcolorbox}[colback=gray!10, colframe=black!50, boxrule=0.5pt, arc=4pt, left=6pt, right=6pt, top=6pt, bottom=6pt]
\textbf{Example Predefined Prompt:}

    \{
        "system": "Your task is to match the user's query to the most relevant intent from a given list. If the current intent sufficiently addresses the query, do not change the intent. Ensure the selected intent exactly matches one from the list, considering case sensitivity.\\

Intents: \{all\_intents\}\\

Query: \{query\}\\

Current Intent: \{current\_intent\}\\

Output only a JSON blob with the key "intent" and no preamble or explanation.\\
        
        "human": ""
    \}
\end{tcolorbox}

\subsection{Intent Recognition and Slot Mapping}
Once a query is classified as ‘Domain Specific’, it is forwarded to the intent recognition module. This module utilizes a predefined set of intents tailored to specific crops and agricultural domains. The module leverages the Mistral model used in the Classifier Module to classify the query into one of these intents. The model analyzes the semantic content of the query, identifies the user's primary goal, and matches it with the closest intent from the predefined intent bank.
Upon identifying the intent, the Mistral model is again utilized to extract the associated slot values that are crucial to understanding the query context. Each intent has 2-5 slots, which include values such as state of cultivation, seed variety, season and other crop specific parameters. In cases where the extracted slot values are incomplete, ambiguous, or missing, the slot mapping module comes into play. This module ensures that all necessary slots are populated by interacting with the user through a dynamic clarification process using the Mistral model. The model generates questions for the required slot values and presents them in the user’s preferred modality - text or speech. As the user responds, the system parses the answers, extracts the required values and populates the missing slots, progressively forming a semantic representation of the query. The iterative slot filling process typically takes 2-3 dialogue turns to complete.

\begin{tcolorbox}[colback=gray!10, colframe=black!50, boxrule=0.5pt, arc=4pt, left=6pt, right=6pt, top=6pt, bottom=6pt]
\begin{center}
    \includegraphics[width=\textwidth, keepaspectratio]{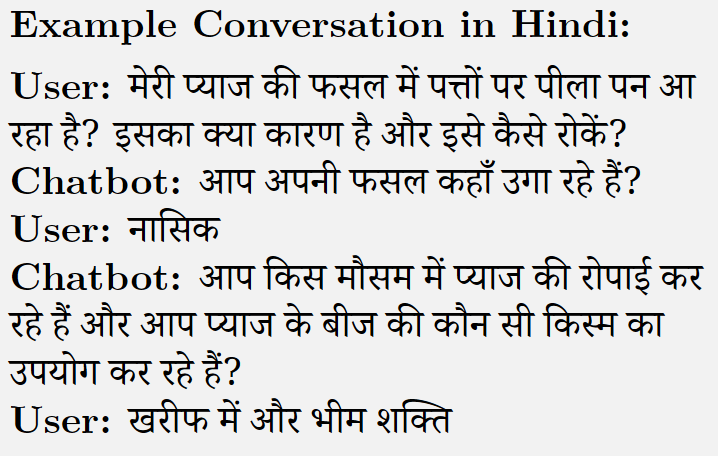}
\end{center}
\end{tcolorbox}

\subsection{Context Retrieval, Answer Generation and User Feedback Loop}

Once the query is enriched, it is contextualized using the retrieval module mentioned in Section 3.5, after which it is given as input to a domain specialized LLM (Param), which has been fine-tuned to align with domain specific knowledge as specified in Section 3.6. The model employs in context learning (ICL) by using a few-shot prompt - a structured template that includes 2-3 curated examples of similar queries and their ideal responses. The user query is fit into this template dynamically along with the retrieved knowledge passage from the retrieval module. The complete prompts span approximately 100 to 200 tokens, which, even after combined with context, is well within Param’s context length of 2048 tokens. This conditioning enables the model to generate a precise, grounded, and fluent response without making modifications to the model weights. The generated output is then formatted and delivered to the user in their chosen modality (speech or text).
\section{Discussion}
The proposed speech enabled, multilingual, AI powered chatbot platform represents a significant step in transforming the agricultural advisory system of India. Designed while keeping in mind the diverse target audience and the need for specialized guidance across the country, the system addresses longstanding challenges such as language barriers, illiteracy, limited access to expert consultation, and the lack of personalized data driven recommendations.

\begin{tcolorbox}[colback=gray!10, colframe=black!50, boxrule=0.5pt, arc=4pt, left=6pt, right=6pt, top=6pt, bottom=6pt]
\begin{center}
    \includegraphics[width=\textwidth, keepaspectratio]{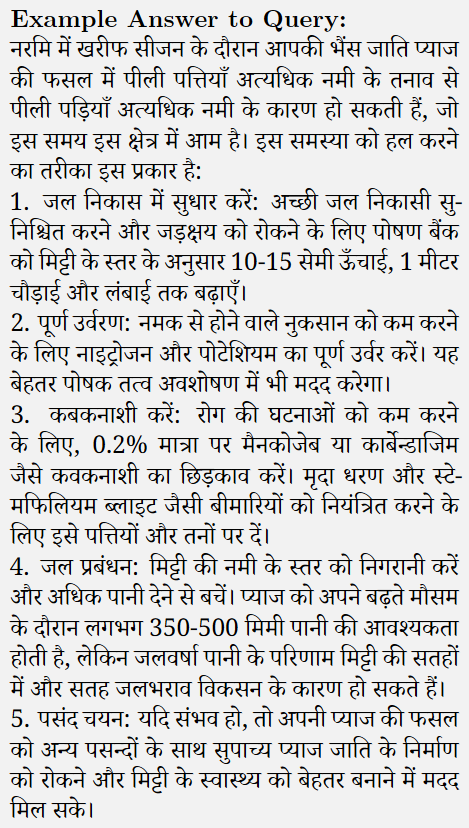}
\end{center}
\end{tcolorbox}
A major strength of the platform lies in its interactive design. Unlike existing static advisories that rely on predefined responses, our solution has the ability to dynamically engage with the user, enabling tailored responses through increased contextual understanding. The conversation interface is accessible via both speech and text, and inculcates both English and Hindi, two of the most widely spoken languages in India. This user interface mimics the experience of interacting with a real life expert, which will build trust among users and increase its adoption among digitally less literate farmers.

The technical architecture of the proposed system plays a pivotal role in enabling this interactivity. By employing a RAG approach with over 150,000 documents backed by a domain specialized Param model using IFT, the pipeline ensures that the final answers are practically applicable with more confidence than existing static systems. The inference functions across the pipeline utilize in context learning through few-shot learning prompts, further enhancing the quality of the answer. This configuration enables improved generalization to unseen queries while avoiding direct modification of model weights.

While the platform shows strong potential, several limitations and challenges remain. The model is currently specialized in two crops: onions and grapes. This approach was taken as an experiment to evaluate the performance of the pipeline on a limited number of crops before extending it to a wider variety. Since domain specialization is achieved through intent-slot mapping and the RAG module, support for additional crops can be scaled by expanding the existing dataset used to build the indexed vector database. Further, expanding multilingual support to include more than 22 languages of India would significantly enhance inclusivity. This can be integrated into our pipeline using specialized translators for each model, along with a finer language detection model.

From an ethical and deployment perspective, the system minimizes hallucination by grounding responses in expert-verified knowledge. As it scales, challenges such as data privacy, fairness, and cultural sensitivity must be continuously addressed. Ensuring secure handling of user data and avoiding regionally inappropriate or biased advice will be critical for responsible and trustworthy deployment.

Looking forward, several enhancements are envisioned: integration of a wider range of crops and languages to increase inclusivity, as mentioned above, integration with IoT sensors and government databases for automated field data collection, and interoperability with national agricultural platforms to expand reach. 

Ultimately, the system lays the groundwork for scalable, AI-driven agricultural support that is accessible, reliable, and tailored to farmers’ real world needs. Its modular design ensures adaptability to diverse crops, languages, and evolving field conditions, making it easy to extend, customize, and integrate with future technological and domain-specific advancements.

\begin{table*}
  \centering
  \resizebox{\textwidth}{!}{%
  \begin{tabular}{lll}
    \hline
    \textbf{Metric Category} & \textbf{Metric} & \textbf{Score / Observation} \\
    \hline
    \multirow{4}{*}{Query Accuracy \& Quality}
      & Query Response Accuracy (QRA) & 97.53\% \\
      & Personalization \& Contextual Relevance (PCR) & 91.35\% (Context-aware and personalized) \\
      & Follow-up Question Relevance (FQR) & 82.85\% (Relevant or appropriately absent) \\
      & Escalation Accuracy (EA) & Not triggered in most cases \\
    \hline
    \multirow{5}{*}{Usability}
      & Average Response Time (RT) & $\sim$5.96 seconds \\
      & Query Completion Rate (QCR) & 97.53\% for test cases \\
      & Speech Input Recognition Accuracy (SIRA) & Partial data; preliminary results promising \\
      & Multimodal Output (TEXT) Delivery Rate & 98\% (All text outputs delivered successfully) \\
      & Multimodal Output (VOICE) Delivery Rate & Partial data; early performance satisfactory \\
    \hline
    \multirow{3}{*}{User Satisfaction}
      & User Satisfaction Score (USS) & High (via feedback) \\
      & Net Promoter Score (NPS) & Positive (Users likely to recommend) \\
      & Helpfulness Score (HS) & High (Rated top-end on scale) \\
    \hline
    \multirow{3}{*}{System Efficiency \& Reliability}
      & Uptime Rate (UR) & $\sim$100\% during testing \\
      & Error Rate (ER) & Negligible \\
      & Escalation Response Time (ERT) & Within acceptable limits \\
    \hline
    \multirow{3}{*}{Language Performance}
      & Language Input Accuracy (LIA) & High (Both Hindi and English) \\
      & Language Output Accuracy (LOA) & High (Both Hindi and English) \\
      & Language-Specific Satisfaction (LSS) & High (Uniform satisfaction across languages) \\
    \hline
  \end{tabular}%
  }
  \caption{\label{evaluation-metrics}
    Summary of evaluation metrics for the agricultural chatbot. The model demonstrates strong performance in accuracy, usability, and multilingual capabilities, with consistently high satisfaction across diverse interaction modalities.
  }
\end{table*}

\section{Results and Conclusion}

\begin{figure}[h]
    \centering
    \fbox{%
        \includegraphics[width=0.8\linewidth, keepaspectratio]{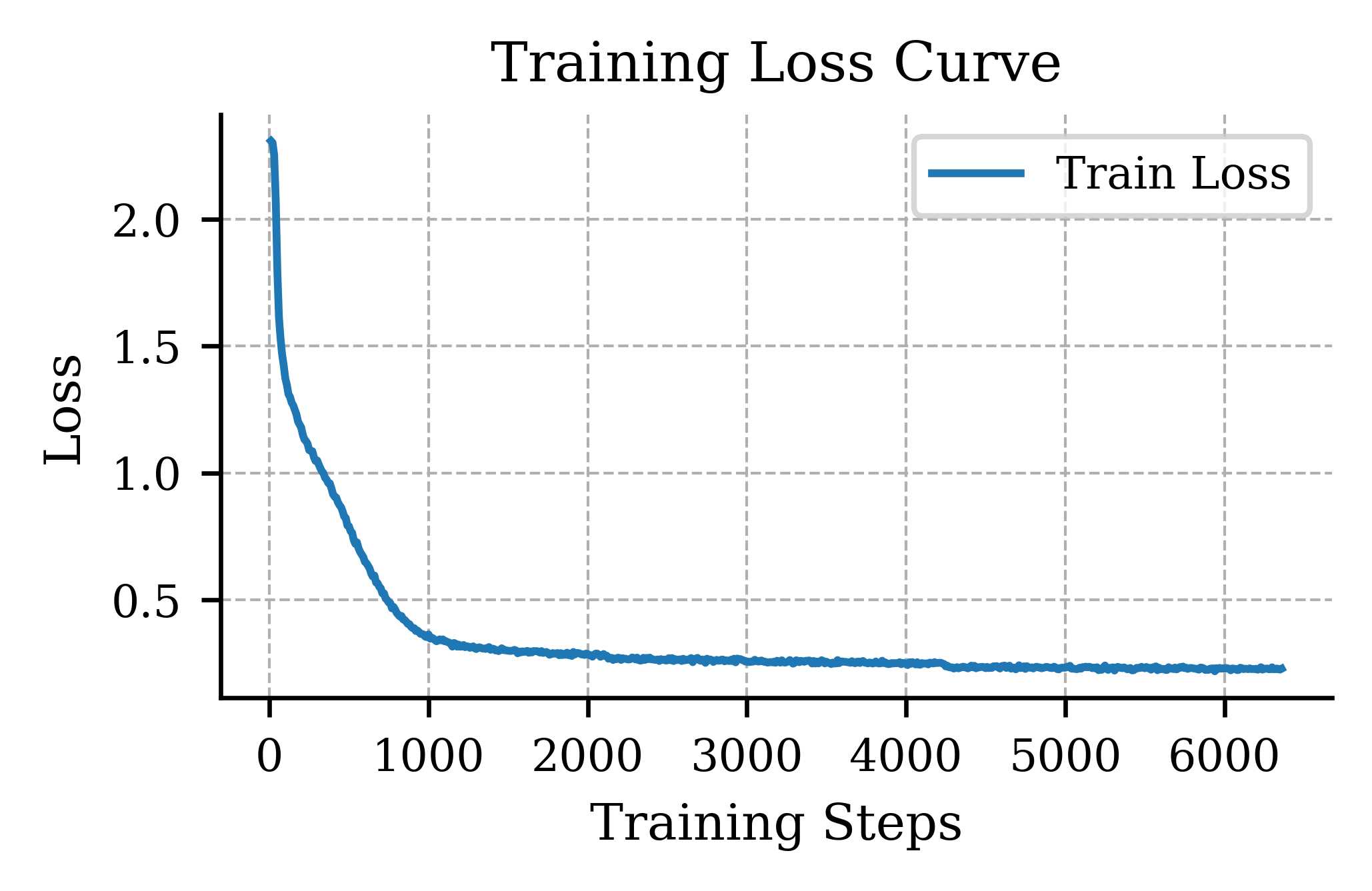}
    }
    \caption{Training Loss Plot}
    \label{fig:train_loss}
\end{figure}

\begin{figure}[h]
    \centering
    \fbox{%
        \includegraphics[width=0.8\linewidth, keepaspectratio]{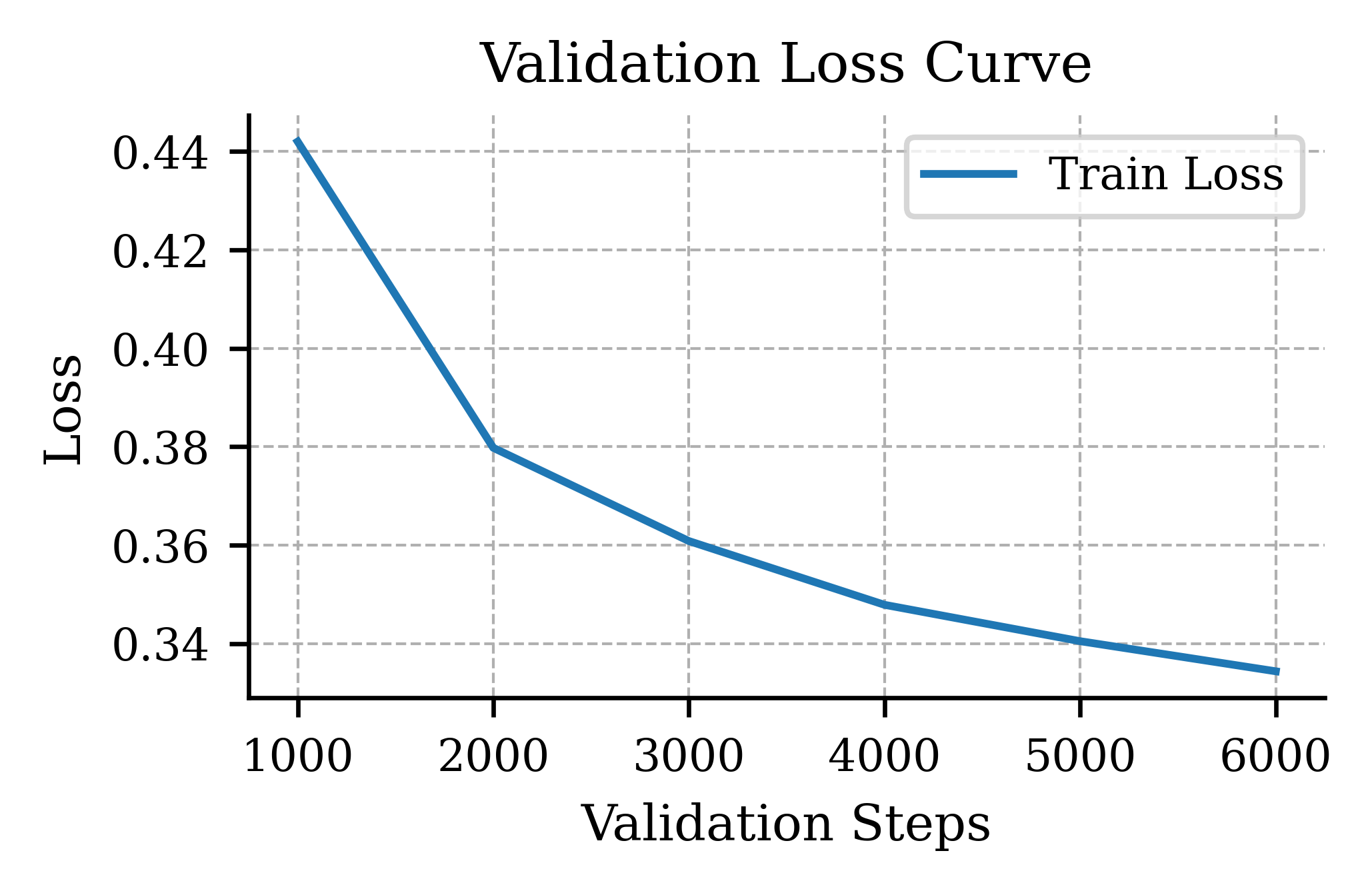}
    }
    \caption{Validation Loss Plot}
    \label{fig:eval_loss}
\end{figure}

The fine tuned model showed strong learning behavior during training, with the loss steadily decreasing from 2.3049 to 0.2239, and the evaluation loss reaching 0.3343, reflecting effective convergence as shown in Figure~\ref{fig:train_loss} and Figure~\ref{fig:eval_loss}. This translated to a strong performance across all key metrics, as shown in Table~\ref{evaluation-metrics}. The Query Response Accuracy (QRA) was 97.53\%, with Personalization and Contextual Relevance (PCR) at 91.35\%, and Follow Up Question Relevance at 82.85\%. These metrics reflect the system's ability to deliver accurate, relevant, and context aware responses. 

In terms of usability, the bot maintained a Query Completion Rate (QCR) of 97.53\% and delivered text output successfully in 98\% cases as well. The average response time was 5.96 seconds, with 53.66\% of responses taking more than 3 seconds, suggesting room for improvement in speed.
User feedback indicates high satisfaction, with strong USS, NPS, and HS scores. The system maintained ~100\% uptime and a negligible error rate throughout testing. Importantly, it performed consistently well in both Hindi and English, with high input/output accuracy and satisfaction across languages.

The evaluation portrays that the proposed chatbot delivers accurate, context aware and user friendly responses, making it suitable for agricultural support, especially in rural and multilingual contexts. Its strong performance in query accuracy, completion, and language and speech handling highlights its real world potential. With minor improvements in response speed, the chatbot is well positioned to support agricultural stakeholders through accessible and intelligent conversations.

\section{Future Work}
\subsection{Image Integration with i-SARATHI-Based Mobile App}
Future iterations of this system will incorporate image analysis capabilities via the i-SARATHI platform. This will enable users to upload images of affected crops or diseases directly through the mobile app, allowing the system to provide visual diagnostic support. Integrating computer vision models will improve disease identification and enhance the system’s diagnostic precision.
\subsection{On Demand In Field Soil Testing via SAMBHAV}
Integrating the current system with SAMBHAV will allow access the on demand sil testing. This will facilitate soil condition based advisory and personalized nutrient management recommendations. It will also ensure that human errors while observing and presenting the information regarding the soil condition are eradicated since the monitoring will be automated and not require human intervention.
\subsection{Multimodal Parameter Integration}
A key future enhancement involves the integration of diverse agro-environmental parameters, including soil composition, leaf characteristics, and microclimate data (eg temperature, humidity etc) by leveraging sensor based data from systems developed by TIH-IoT.
\subsection{Interoperability with Government Platforms}
Another attractive future enhancement is achieving seamless interoperability with existing government agricultural platforms such as the mKisan portal and the KISAAN 2.0 mobile application. This will ensure uniformity in communication, better outreach, and a centralized repository of farmer interactions and advisory records.
\subsection{Multilingual Support Across Indic Languages}
The system will be scaled to provide support in 22 official indic languages, ensuring accessibility for farmers across different linguistic backgrounds. This will involve developing multilingual pipelines across both the text and speech domain.

% Bibliography
\bibliographystyle{apalike}
\bibliography{custom}

\end{document}